\documentclass[Runningheads]{llncs}
\usepackage[T1]{fontenc}
\usepackage{graphicx}
\usepackage{float}
\usepackage{algorithm}
\usepackage{algorithmicx}
\usepackage[noend]{algpseudocode}

\usepackage[title]{appendix}
\usepackage[marginal]{footmisc}

\usepackage{color}
\usepackage{subfigure}

\begin{document}
\title{Reflection of Episodes: Learning to Play Game from Expert and Self Experiences}
%
%
\author{
Xiaojie Xu\inst{1} \and
Zongyuan Li\inst{1} \and
Chang Lu\inst{1} \and
Runnan Qi\inst{2} \and
Yanan Ni\inst{2} \and
Lumin Jiang\inst{2} \and
Xiangbei Liu\inst{1} \and
Xuebo Zhang\inst{1} \and
Yongchun Fang\inst{1} \and
Kuihua Huang\inst{2,*} \and
Xian Guo\inst{1,*} \and
Zhanghua Wu\inst{3} \and
Zhenya Li\inst{4}
}
\authorrunning{Xiaojie Xu et al.}
%
\institute{College of Artificial Intelligence, Nankai University, Tianjing, China \\
\email{\{2120220490, 2120230524, 2111533, 2120230527\}@mail.nankai.edu.cn}\\
\email{\{fangyc, zhangxuebo, guoxian\}@nankai.edu.cn}
\\
\and
Laboratory for Big Data and Decision, National University of Defense Technology, Changsha, China\\
\email{\{qirunnan13579, niyanan, jlm\_mz, khhuang\}@nudt.edu.cn}\\
\and
Jiangsu Automation Research Institute Jiangsu, China\\
\email{wuzhanghuamaoli@163.com}
\and
Nanjing Research Institute of Electronic Engineering\\
}
\maketitle              
\footnote{* Corresponding author.}
\vspace{-0.5cm}
\begin{abstract}
StarCraft II is a complex and dynamic real-time strategy (RTS) game environment, which is very suitable for artificial intelligence and reinforcement learning research. To address the problem of Large Language Model(LLM) learning in complex environments through self-reflection, we propose a Reflection of Episodes(ROE) framework based on expert experience and self-experience. This framework first obtains key information in the game through a keyframe selection method, then makes decisions based on expert experience and self-experience. After a game is completed, it reflects on the previous experience to obtain new self-experience. Finally, in the experiment, our method beat the robot under the Very Hard difficulty in TextStarCraft II. We analyze the data of the LLM in the process of the game in detail, verified its effectiveness.
\end{abstract}
\section{Introduction}
StarCraft II is a complex and dynamic real-time strategy (RTS) game environment, which is very suitable for artificial intelligence and reinforcement learning research. In "StarCraft II," players choose a race in the game and defeat opponents through resource management, base construction, technology upgrades, and combat. The game has a high level of strategic and tactical depth, requiring players to excel in economic management, army control, and decision-making. In 2017, DeepMind and Blizzard Entertainment collaborated to release SC2LE\cite{SC2LE}, a standardized reinforcement learning environment aimed at promoting AI research. SC2LE provides an API that allows researchers to control units and buildings in the game and obtain game state information. In 2019, DeepMind released AlphaStar\cite{AlphaStar}, an AI system that uses multi-agent reinforcement learning and can compete with top human players in "StarCraft II." Reinforcement learning and other methods demonstrate strong operational capabilities in the "StarCraft II" environment, but at the same time, these agents most just focus on the details of the current state, but ignore the overall strategy.

\begin{figure}
\centerline{\includegraphics[width=360pt]{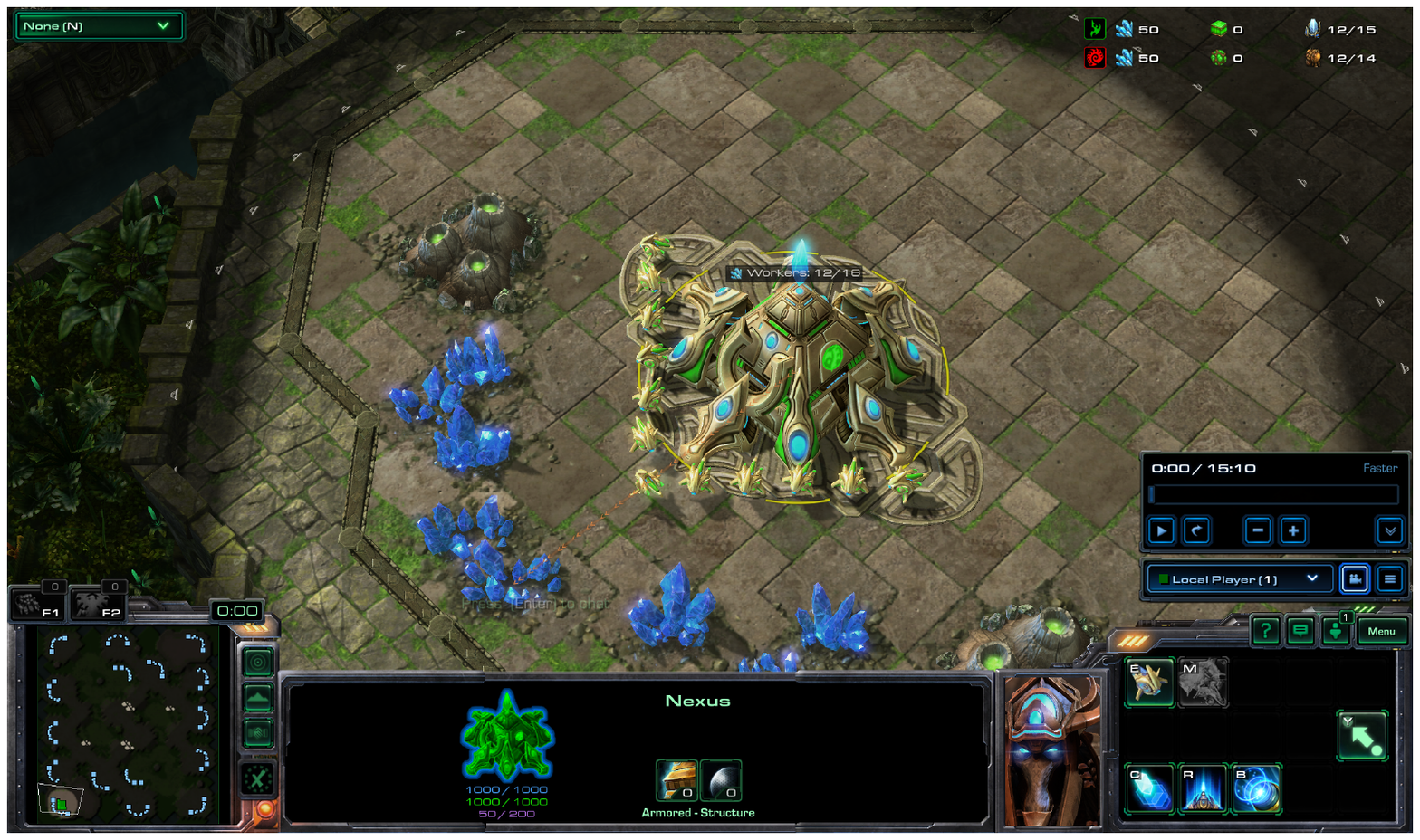}}
\caption{\textbf{StarCraft II.} A complex and dynamic real-time strategy game environment, which make it very suitable for artificial intelligence research.} 
\end{figure}

In June 2018, GPT-1 was released, demonstrating the potential of large language models in natural language understanding and generation tasks. In November 2022, OpenAI released ChatGPT\cite{ChatGPT}, which can interact based on the context of the conversation, truly chat and communicate like a human, and answer questions. April 2023, OpenAI released GPT-4\cite{GPT-4}. Building on these, a team from the Chinese Academy of Sciences constructed a text interaction environment called TextStarCraft II\cite{Baseline}, allowing large models to access various data in the environment, formulate strategies, and control agents in "StarCraft II" through commands. In their paper, they proposed a way based on  Chain of Summary(COS) to enable large models to make decisions, defeating robots at a difficult level. However, this method has some flaws: The decision-making information in the game is redundant; the game process lacks reflection, which hinders the improvement of gaming skills.

In this paper, we propose Reflection of Episodes(ROE) framework based on expert experience and self-experience. This framework obtains key information in the game through a keyframe selection method divided into stages, and then makes decisions based on expert experience and self-experience. After the game is completed, it reflects on the previous experience to obtain new self-experience. In this way, LLM can achieve strategy iteration and continuous improvement in reflection. In the experiment, our method beat the robot under the Very Hard difficulty in TextStarCraft II, verified its effectiveness.
\section{Related Works}
\subsection{StarCraft-II Environment}
Prior to the application of AI to StarCraft II, AI research in RTS games had been going on for years. Early attempts focused on using rules and finite state machines to create simple game agents. These agents are often inflexible and adaptable, unable to handle the complex strategies of human players. 

In 2017, DeepMind and Blizzard Entertainment collaborated to release SC2L
E\cite{SC2LE}. It is a standardized reinforcement learning environment designed to facilitate AI research. SC2LE provides apis that allow researchers to control units and buildings in the game and access game state information. SC2LE provides a unified test bed for researchers, greatly advancing StarCraft II AI research. 

In 2019, DeepMind released AlphaStar, a landmark AI system that uses Multi-Agent Reinforcement Learning to compete against top human players in StarCraft II. In multi-agent reinforcement learning, there are also many methods used in StarCraft II environments, such as VDN\cite{VDN}, QMIX\cite{QMIX}, WQMIX\cite{WQMIX}, MAPPO\cite{MAPPO}, MADDPG\cite{MADDPG}. In recent years, there have been other studies of the StarCraft II environment through other methods, such as RLforSC\cite{RLforSC}.

However, existing methods cannot improve the interpretability of AI decision-making processes or help researchers understand and improve AI strategies.

\subsection{Large Language Model}
LLMS typically have complex architectures with billions or even hundreds of billions of parameters that are trained with vast amounts of data and computational resources. In June 2018, GPT-1 was the first generative pre-trained converter model released by OpenAI. It uses the converter architecture for unsupervised pre-training on large amounts of text data and then optimization on specific tasks through supervised fine-tuning. In November 2022, ChatGPT was released as a chat application. 

In recent years, there have also been a variety of large models in different directions, such as Claude-2\cite{Claude-2}, LLAMA-3\cite{LLAMA-3}, PaLM-2\cite{PaLM-2} and so on. Claude is a large-scale language model introduced by Anthropic that focuses on security and controllability to reduce harmful information in generated content. LLaMA-3 is a large-scale language model published by Meta to advance AI research and provide an open research platform. 

The development of large language models has also facilitated the study of StarCraft II environments, such as SwarmBrain\cite{SwarmBrain}, but these studies have not solved the problem of reflection and strategy iteration in complex environments with LLMs.

\subsection{LLM and reflection}
In order to make the large model continuously improve itself in the process of communication, the method of combining the large model and reflection is constantly proposed. A new framework, reflexition\cite{reflextion}, was proposed in 2023, which instead of updating weights, strengthens language agents through language feedback, allowing large models to learn quickly and efficiently from trial and error.In this approach, the focus is on action-level reflection, allowing the large model to store buffers in each chat, thereby inducing better decisions in subsequent exchanges.
In addition to action-level reflection, Agent-Pro\cite{Agent-Pro} methods are proposed to form strategy-level reflection and optimization, fine-tuning its overall strategy by reflecting on past trajectives and decisions.

However, these reflection methods do not take into account the adaptability of frameworks in complex environments such as StarCraft II. Therefore, we propose Reflection of Episodes to realize strategy-level reflection in complex environment.

\begin{figure}[htbp]
\centerline{\includegraphics[width=360pt]{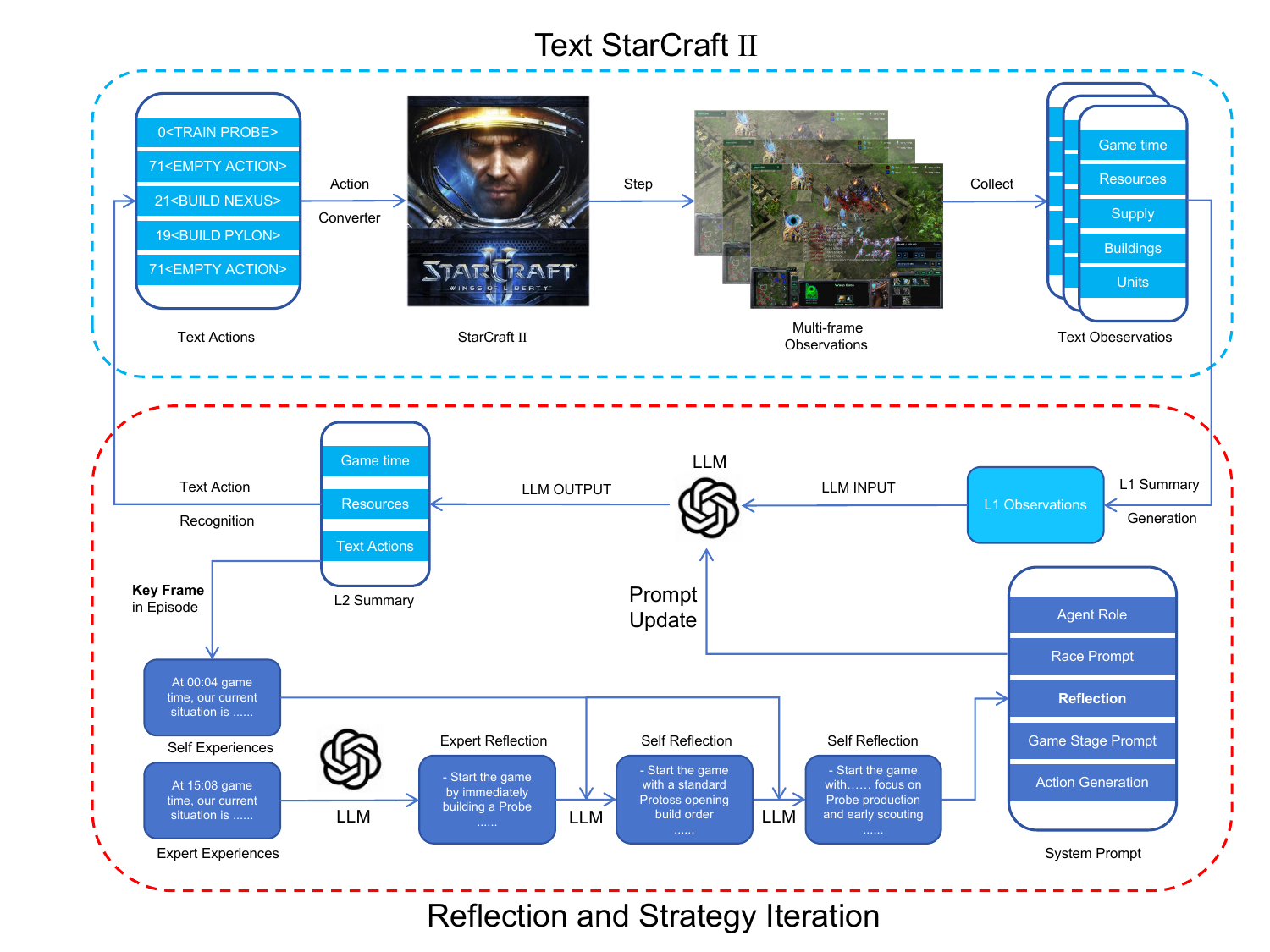}}
\caption{\textbf{Reflection of Episodes Framework.} The framework consists of Text StarCraft 2 environment and reflection structure. After an episode, reflection structure generate new prompt and update it to the next game.} \label{fig2}
\end{figure}

\section{Methods}

We propose a framework for reflection and strategy iteration as Figure 2. The ROE framework mainly includes two key improvements: key frame selection based on game phase division; reflection generation and strategy iteration based on reflection of episodes. All the prompts used are shown in Appendix A.

\subsection{Key Frame Selection}

In an entire StarCraft II match, the environment will generate 7000 frames of data for about 20 minutes, and the large model will interact with the environment more than 700 times. In the process of reflection generation, in order to summarize the whole game situation with as little data as possible, we proposed a key frame selection method based on game phase division.

\begin{algorithm}
    \caption{Key Frame Selection}
    \begin{algorithmic}[1]
    \Require L2 summary of an Episode: $L2\_summary$.
    \Ensure Key Frame in Episode: $key\_frame$
        \State Initilalize $key\_frame$ 
        \State Initilalize $game\_phase\_transision$ 
        \State $data = read\_file(L2\_summary)$
        \For{$frame$ in $data$}
            \If{$game\_phase\_transision$ in $frame$}
                \State $key\_frame.appende(frame)$
            \EndIf
        \EndFor
        \State return $key\_frame$
    \end{algorithmic}
\end{algorithm}

\subsubsection{Game Phase Division}

During the course of StarCraft II, resources, buildings, units, etc. change over battlefield situation, so game stages can be divided by these data. We came up with a standard for dividing StarCraft II game stages and included it in the prompt, so that the LLM includes an analysis of the current phase when it generates the L2 Summary. In the subsequent key frame selection process, the most critical part of the game is selected according to these stage analyses. Detailed prompt is shown in Appendix A. 

\subsubsection{Key Frame Selection}
Through the game phase division prompt, the L2 Summary of the large model output contains game phase keywords, such as "Early Mid Game". In algorithm 1, the key frame of the game stage changes is located by keyword search, and then the data near it is selected as the key frame. In order to ensure the integrity of the data, the average selected frames are added as the final input data of reflection.

\subsection{Reflection Generation and Strategy Iteration}

After the completion of the generation of key frames, a new self-reflection is generated according to the previous experience and the key frame of the game, and then the self-reflection is added to the system prompt to achieve strategy iteration. As shown in Figure 3, in three consecutive games, the strategy based on expert experience and first self-reflection failed against the Very Hard robot. The second self-reflection based strategy, obtained after two iterations, wins the game.
Below we will cover in detail the methods of reflection generation and strategy iteration.

\begin{figure}[htbp]
\centerline{\includegraphics[width=360pt]{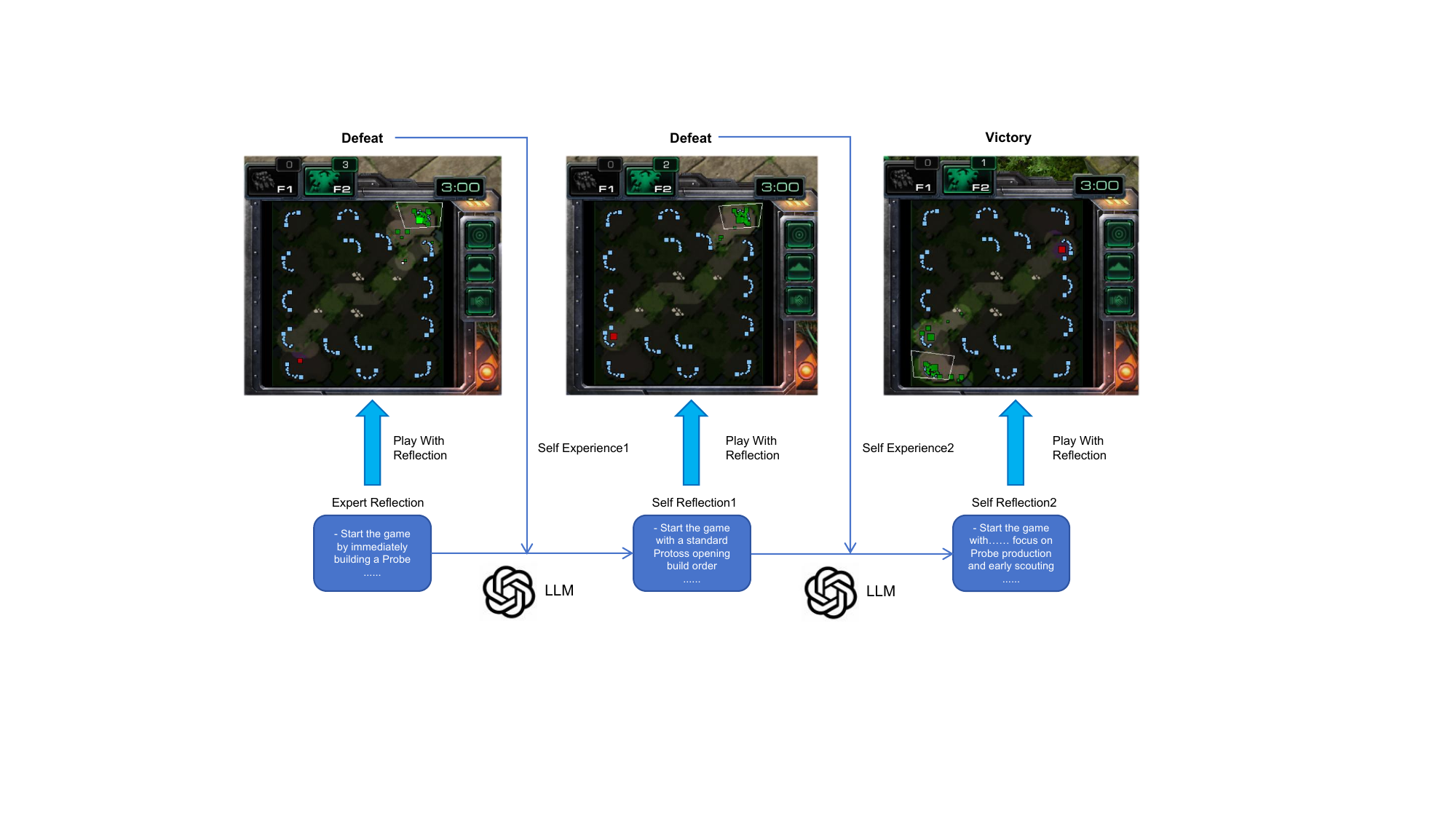}}
\caption{\textbf{Reflection and Strategy Iteration.} In three consecutive games, expert reflection and two generations of self-reflection took the game from defeat to victory.} \label{fig3}
\end{figure}

\begin{algorithm}
\caption{Reflection Generation and Strategy Iteration}
\begin{algorithmic}
\Require L2 summary of an Episode $L2$; Text observation $o_t$; Agent role prompt $p_r$; Expert experience $expert\_experience$; Legal Action Library $L$; Max Round in a test $max\_round$; Reflection Prompt$p\_reflection$.
\State Initilalize environment $Env.reset()$
\While{not $Result="Victory"\ or\ round=max\_round$}
        \State Initilalize L2 summary $L2.reset()$
        \If{$round = 0$}
            \State $self\_experience=expert\_experience$
        \Else
            \State $self\_experience=self\_reflection$
        \EndIf
        \While{not $Env.is\_end()$}
        \State $m_t.reset()$
        \State $m_t.append(p_r+self\_experience+L)$
        \State $a_t = LLM.query(m_t)$
        \State $L2.append(a_t)$
        \State $Env.step(a_t)$
        \EndWhile
        \State $key\_frame=key\_frame\_selection(L2)$
        \State $m_r=(p\_reflection + key\_frame)$
        \State $self\_reflection = LLM.query(m_r)$
        \State $round = round+1$
\EndWhile
\vspace{-0.1cm}

\end{algorithmic}
\end{algorithm}

\subsubsection{Replay analysis method}

To effectively analyze StarCraft replays and learn from them, we propose a replay analysis method, which is divided into the following three stages. First, When watch the Replay, focus on key aspects as "Opening Build Order"(Timing of first buildings), "Economy Management"(Worker production and expansion time), "Scouting"(send probe to gather information) and so on. After reviewing these aspects, make a list of key mistakes and areas for improvement, look for delayed expansion or other points. Finally, based on these findings, make a plan to address these issues in your future games.

\subsubsection{Response standardization} 
 
In replay analysis, we showed how the LLM can analyze the entire game based on key frames in the prompt. After analysis, we designed Response standardization, and added the previous experience and the result of the match as additional prompts to make it better to generate reflections. In the first game, the previous experience is the pre-set expert experience, which is reflected on the successful experience of a difficult game. The specific structure of Response standardization is as follows: Based on the above information, the large model must generate specific suggestions in the format as well as analysis of 8 specific strategic points and no less than 5 key time points and recommendations.

As the final part of the overall Reflection and Strategy Iteration Framework, the generated reflection replaces previous experience and generates new system prompts. The whole process closes the loop and begins the next round of strategy iteration.

\clearpage
\section{Experiments}

All the experiments run on the TextStarCraft-II environment. The baseline is the COS\cite{Baseline} method in TextStarCraft-II environment. To compare with the baseline results, we used exactly the same settings. In the experiment, our method takes 5 consecutive rounds as one test, and winning within the limit of 5 is regarded as Victory.

\subsection{Experiment Settings}

In this section we will introduce some of the necessary setup parts of the experiment, including some important parameters such as player race, opposite race, difficulty and so on. Detailed data are shown in Table 1.

\begin{table}
\caption{\textbf{Experiment Settings}}\label{tab1}
\begin{center}
\vspace{-0.3cm}
\begin{tabular}{p{6cm} p{6cm}}
\hline
 parameter & setting \\
\hline
env type & text \\
map pool & LADDER MAP 2023 \\
player race & Protoss \\
opposite race & Zerg \\
opposite type & build in \\
opposite bot & hydra ling bane bot \\
difficulty & Hard to Elite \\
agent type & chatgpt \\
LLM model name & GPT 3.5 turbo \\
LLM temperature & 0(1 in reflection) \\
Max round & 5 \\

\hline
\end{tabular}
\end{center}
\end{table}

\subsection{Experiment Results }
In this section, we compared the winning rates of our method with the baseline method. In our experiments, agents with GPT-3.5 played Protoss in TextStarCraft-II against Zerg of varying difficulties. As shown in Table~\ref{tab2}, our method increased the winning rate on Hard and Harder difficulties and managed to defeat VeryHard opponents with a 20\% winning rate.

\begin{figure}[htbp]
\centerline{\includegraphics[width=360pt]{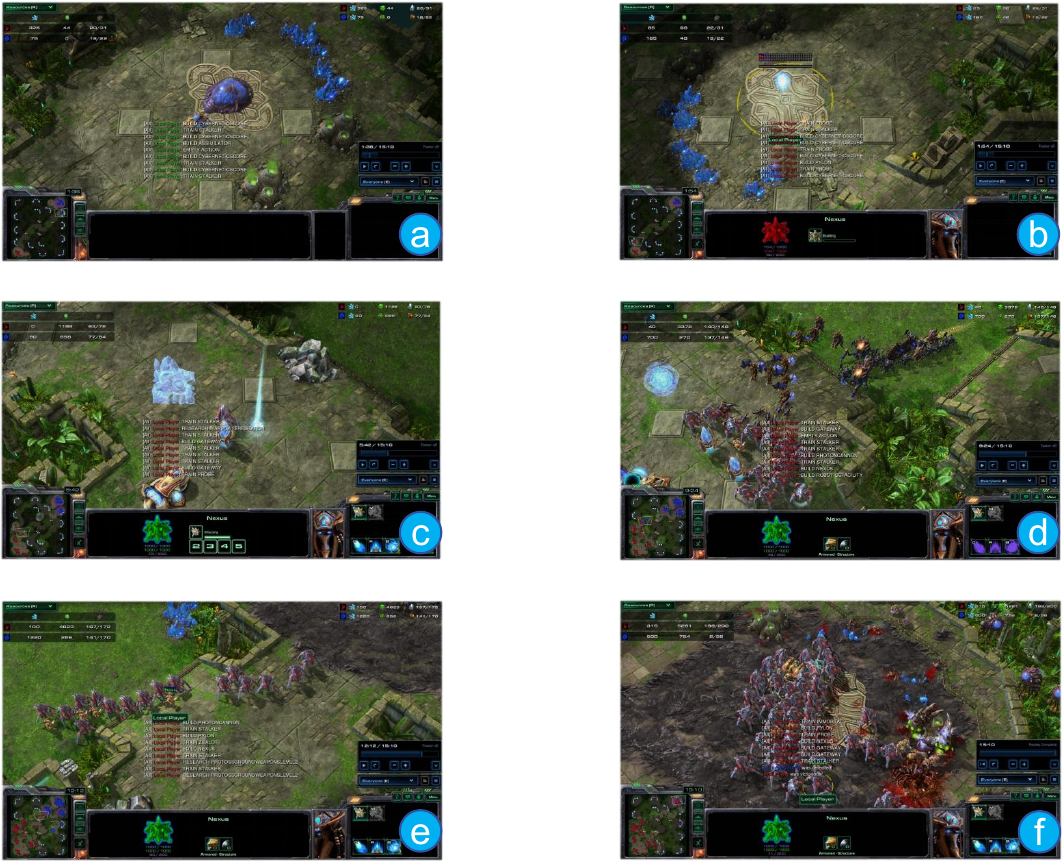}}
\caption{\textbf{Detailed game analysis of Very Hard difficulty} (a) At 1:36 we sent the first reconnaissance plane to detect the enemy; (b) At 1:54 we build our first sub-Nexus; (c) At 5:42 we set up an outpost to respond to the enemy attack; (d) At 9:24, we engaged the enemy in the first major confrontation; (e) At 12 '12 we sent troops to attack the enemy Nexus; (f) At 15:10, we win the game by defeating all enemy Nexus } \label{fig2}
\end{figure}

\begin{table}
\caption{\textbf{Winning Rates in Difficulties Ranging from Level-4 to Level-7.}}\label{tab2}
\begin{center}
\vspace{-0.3cm}
\begin{tabular}{p{2cm} p{2.5cm} p{2.5cm} p{2.5cm} p{2.0cm} }
\hline
Method & \multicolumn{4}{c}{Difficulty Level} \\
\hline
 & Hard & Harder & VeryHard & Elite \\
\hline
Baseline & 21/25 (84\%) & 5/10 (50\%) & 0/12 (0\%) & TBD \\
Ours & 10/10 (100\%) & 10/10 (100\%) & 2/10 (20\%) & 0/6 (0\%) \\
\hline
\end{tabular}
\end{center}
\end{table}

\clearpage
In the experiment, we collected and compared the data of our method with the baseline under Very Hard difficulty, and the data was divided into two aspects, one resource aspect and the other unit aspect, which can best show the situation of the game

\begin{figure}[htbp]
\centerline{\includegraphics[width=360pt]{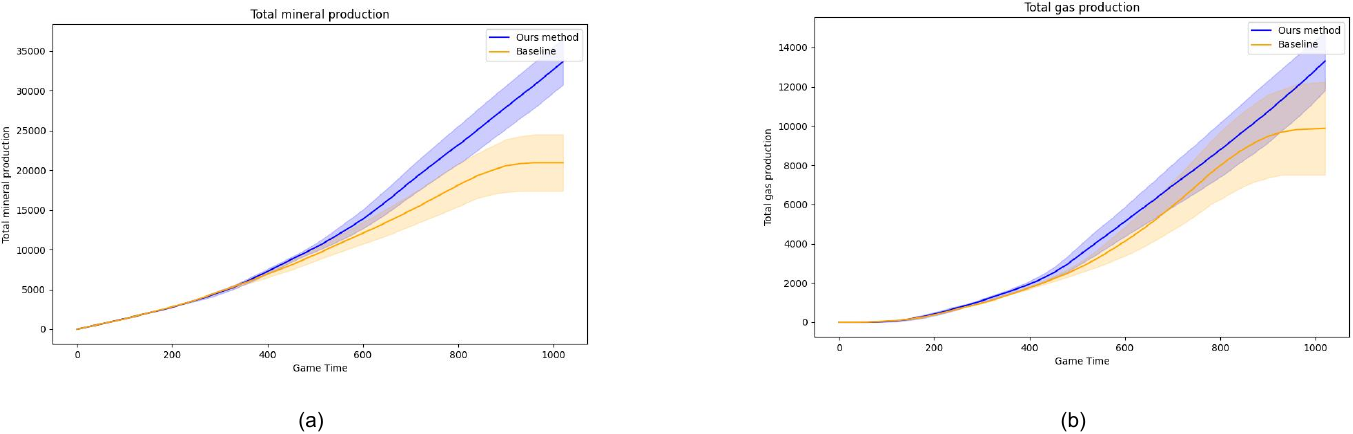}}
\caption{\textbf{Comparison of baseline experiments in resources}} \label{fig2}
\end{figure}

In terms of resources, it can be seen that the total minerals and gases produced by our method have a significant gap in the middle and late period compared with the baseline method, which is one of the reasons why our method can quickly supplement the units in the next unit analysis

\begin{figure}[htbp]
\centerline{\includegraphics[width=360pt]{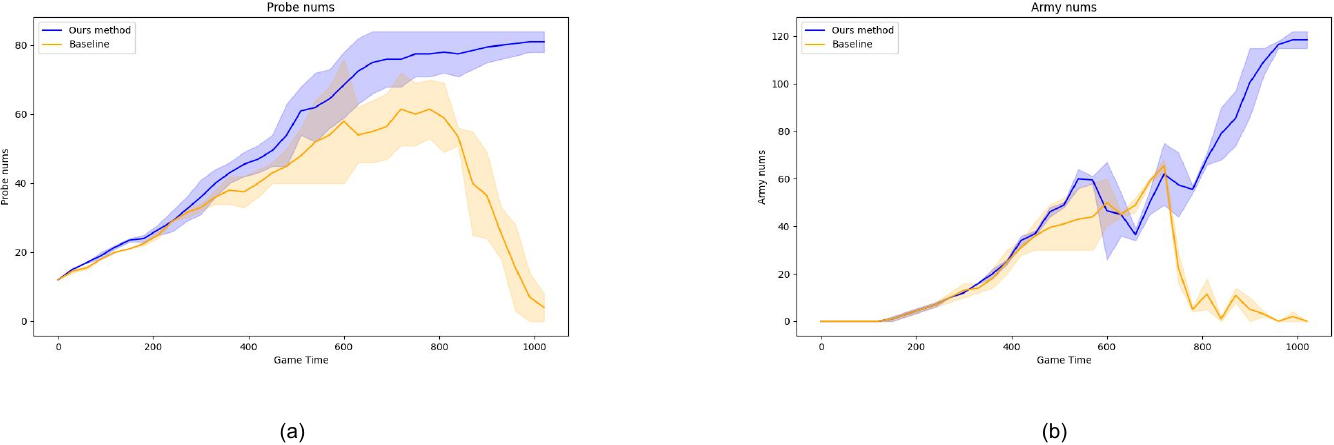}}
\caption{\textbf{Comparison of baseline experiments in units}}\label{fig2}
\end{figure}

On the basis of resources, we look at the comparison between the two sides in terms of units. It can be seen that the baseline probe production rate slowed down in the medium term, and the army could not be effectively replenished after facing a wave of enemy attacks

\subsection{Ablation experiment}
In order to verify the role of keyframe selection, self-reflection and strategy iteration frameworks in the experiment, we conducted an ablation experiment to verify the experimental results and showed their differences by comparing the key data.
\begin{figure}[htbp]
\centerline{\includegraphics[width=360pt]{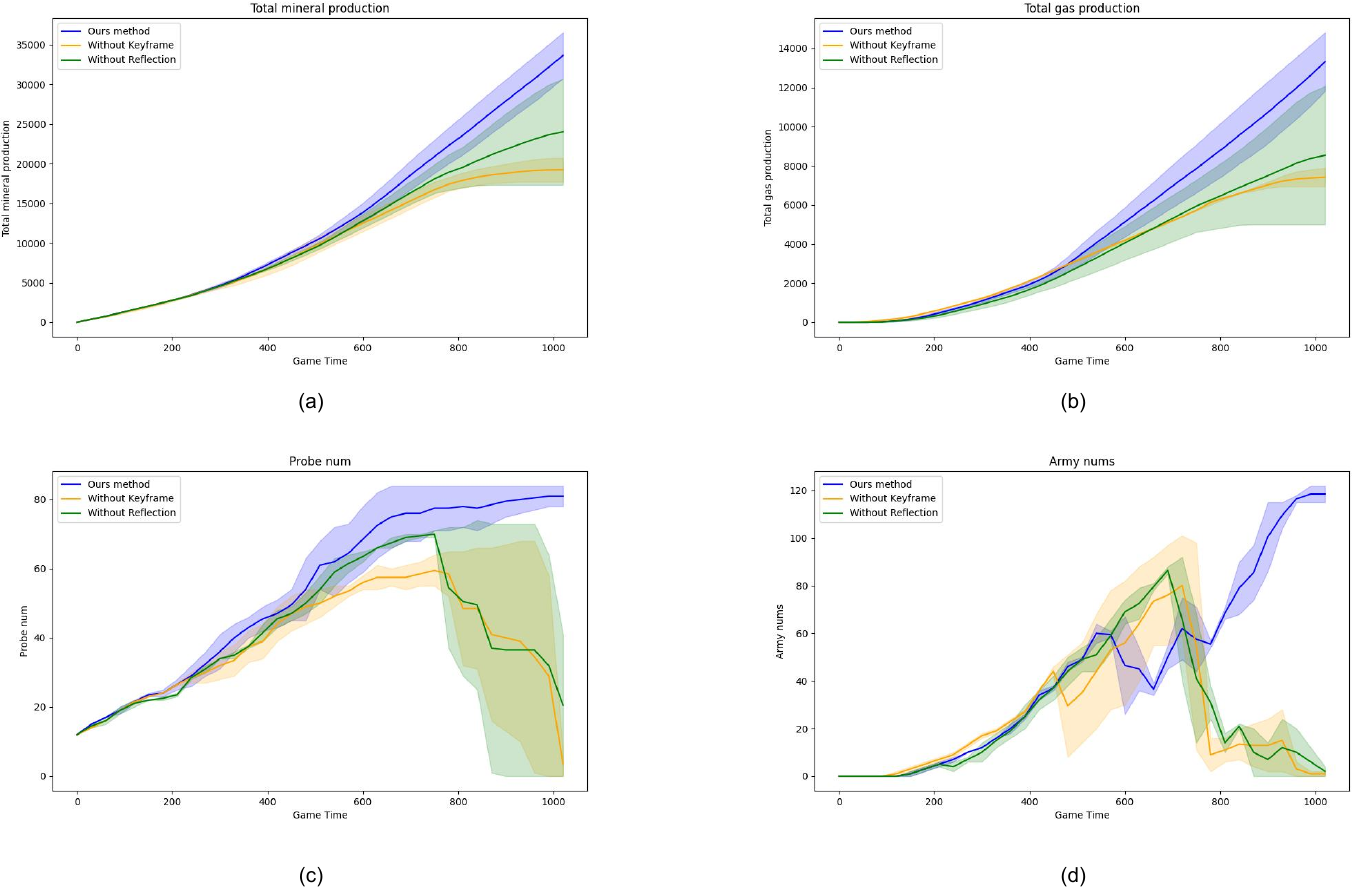}}
\caption{\textbf{Comparison of ablation experiments}} \label{fig2}
\end{figure}

Through ablation experiments, we can find comparison between the two sides in terms of units after losing the method of keyframe selection or reflection. It can be seen that the ablation probe production rate slowed down in the medium term, and the army could not be effectively replenished after facing a wave of enemy attacks.


\section{Conclusion}
In this paper, we propose a ROE framework based on expert experience and self-experience. Through the framework, LLM makes decisions based on expert experience and self-experience. After the game is completed, it obtains key information in the game through a keyframe selection method, then reflects on the previous experience to obtain new self-experience. This framework has been experimentally proven to effectively enhance LLM's strategies through reflections of episodes based on expert experience and self-experience. The reflections obtained also exhibit partial strategy interpretability, which can be regarded as a strategy generation approach using LLMs. Further exploration will be conducted in our subsequent research.

%

\appendix

\clearpage
\section*{Apendix A. All Prompt}
In this appendix, we will show all the prompts used during the experiment, as well as during the experiment, including system prompts, reflection prompts, etc.
\setcounter{figure}{0}
\renewcommand{\thefigure}{A\arabic{figure}}

\subsection*{A.1 System Prompt}
\begin{figure}[htbp]
\centerline{\includegraphics[width=\textwidth]{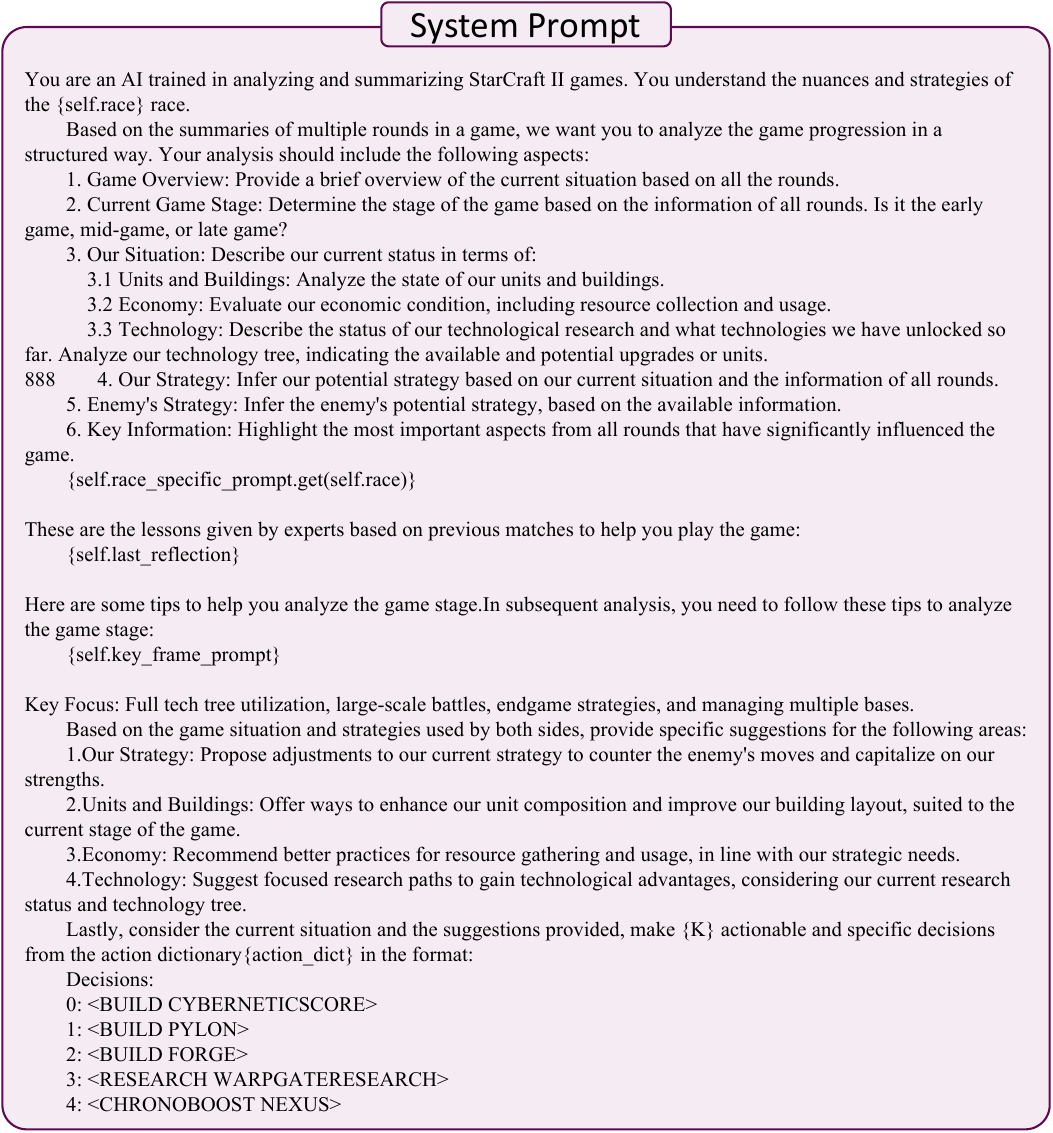}}
\caption{\textbf{General Structure of System Prompt.}} 
\end{figure}

\begin{figure}[htbp]
\centerline{\includegraphics[width=\textwidth]{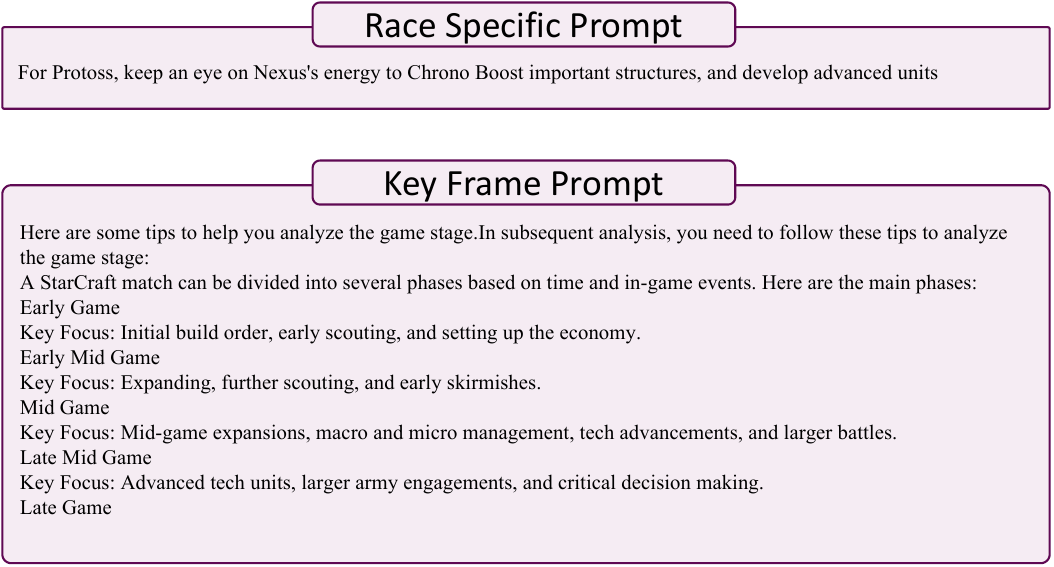}}
\caption{\textbf{Part of System Prompt.} Our Race Specific and Key Frame prompt} 
\end{figure}

\subsection*{A.2 Reflection Prompt}
This section contains the key tips for the reflective section of the entire framework, consisting of 4 main sections: role introduction, Reflective key aspects, task instructions, format requirements.
\begin{figure}
\centerline{\includegraphics[width=\textwidth]{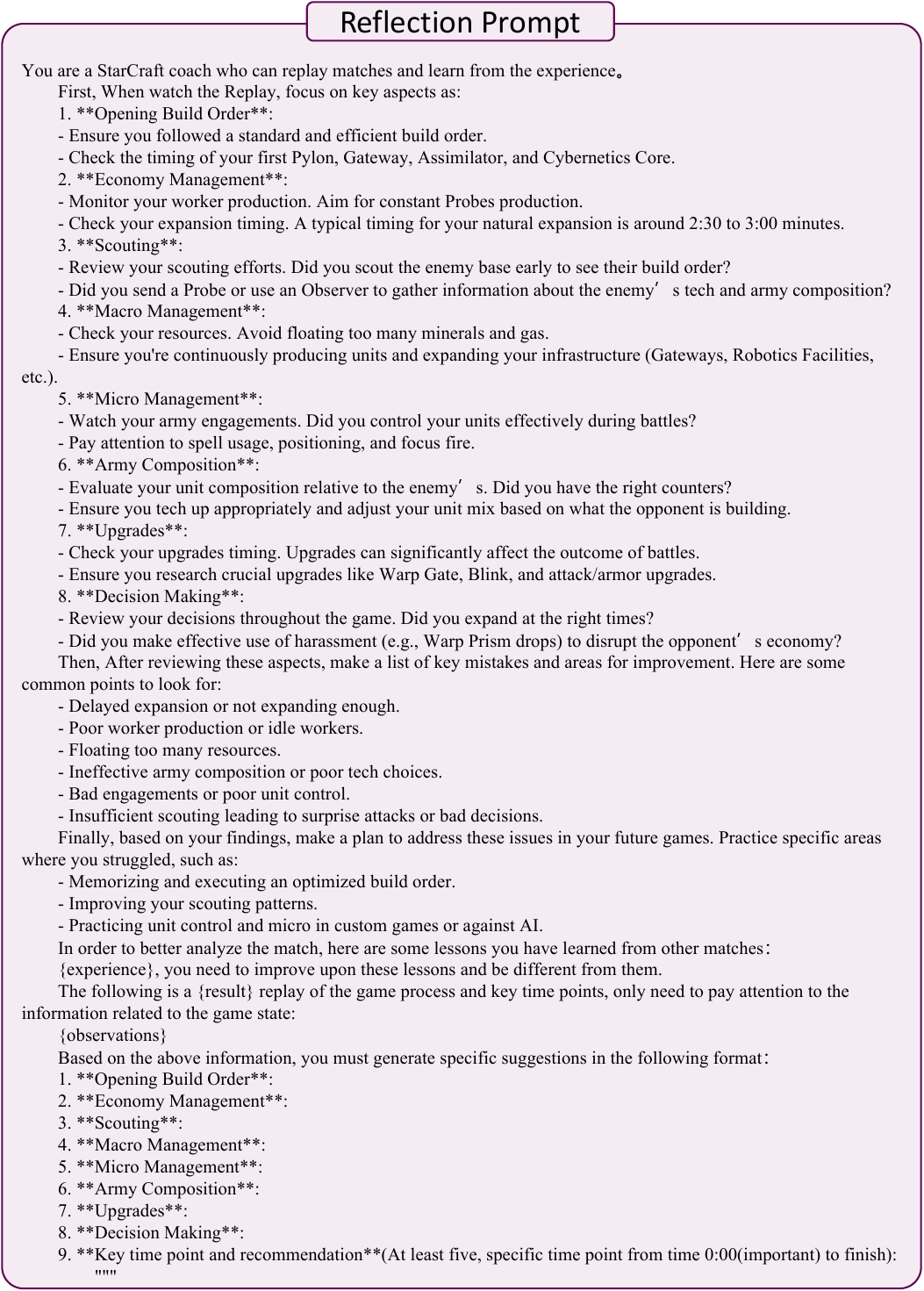}}
\caption{\textbf{Our Reflection prompt.}} 
\end{figure}

\clearpage
\section*{Appendix B. Reflection Iterations}
\setcounter{figure}{0}
\renewcommand{\thefigure}{B\arabic{figure}}
In this appendix, we will show the reflections and changes generated during the experiment of our method under Very Hard built-in AI. The marked part is the part that has changed greatly during the reflection process.

\begin{figure}
\centerline{\includegraphics[width=\textwidth]{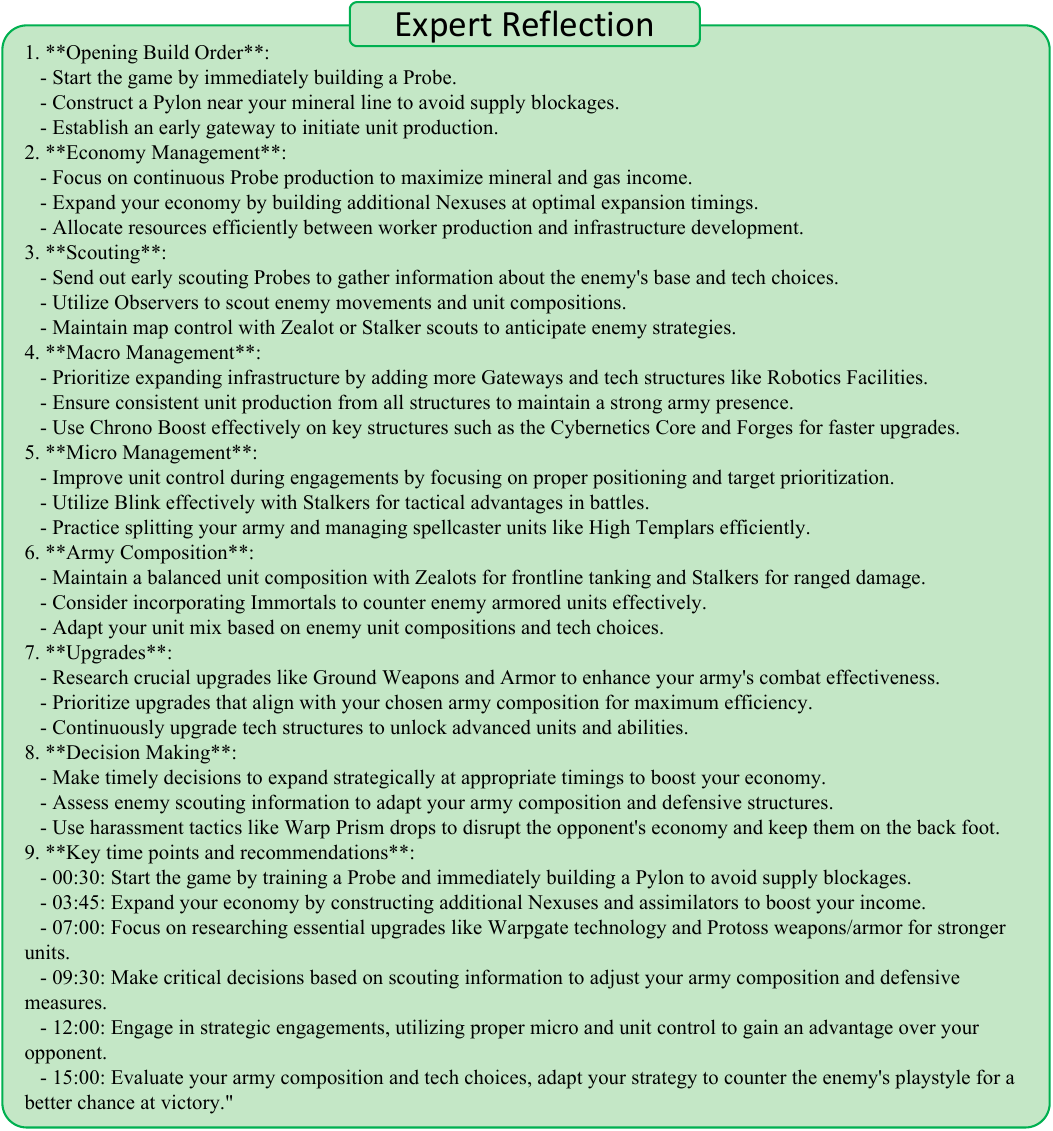}}
\caption{\textbf{Expert Reflection with our method against Very Hard built-in AI}} 
\end{figure}

\begin{figure}
\centerline{\includegraphics[width=\textwidth]{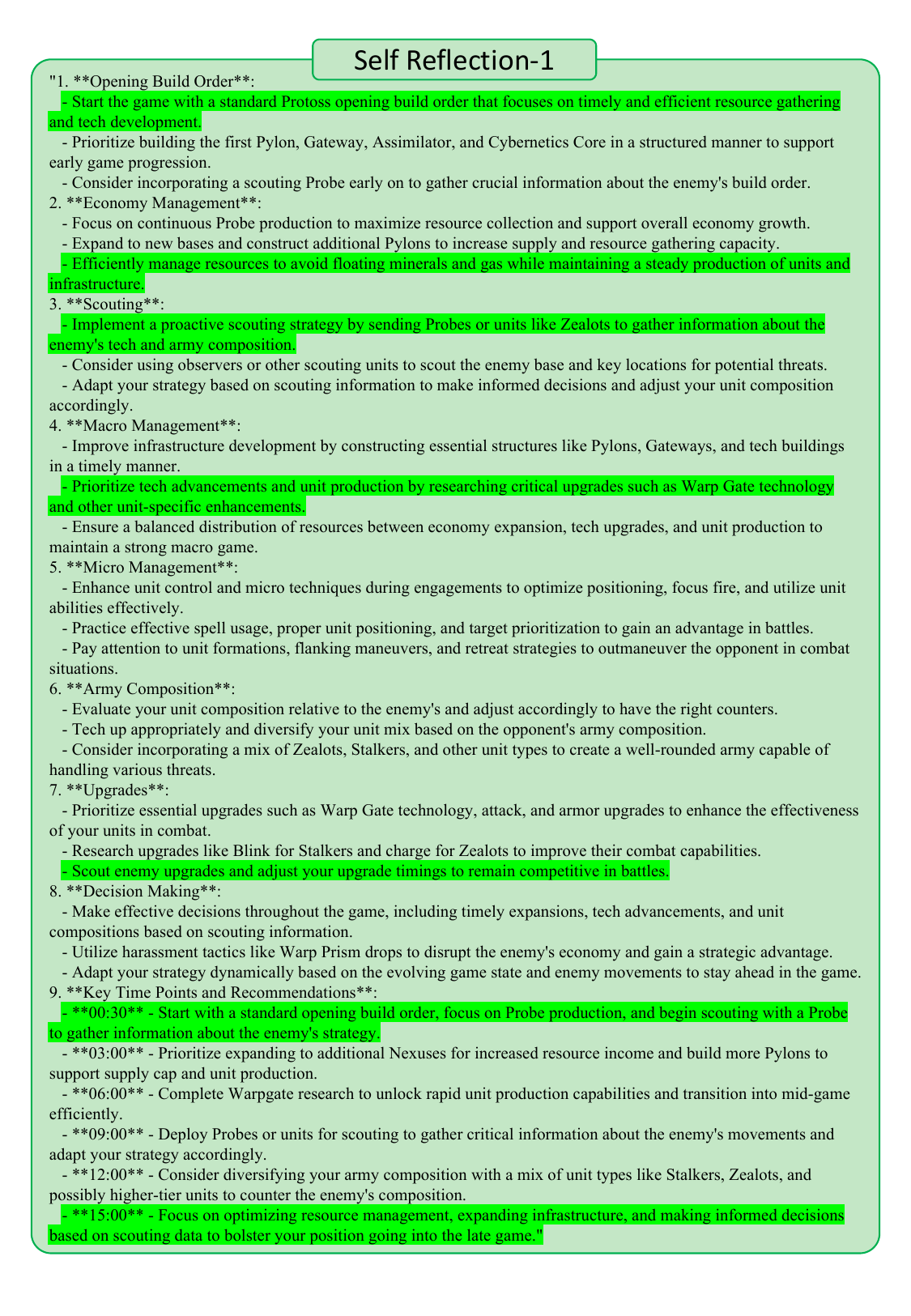}}
\caption{\textbf{Self Reflection1 with our method against Very Hard built-in AI}} 
\end{figure}

\begin{figure}
\centerline{\includegraphics[width=\textwidth]{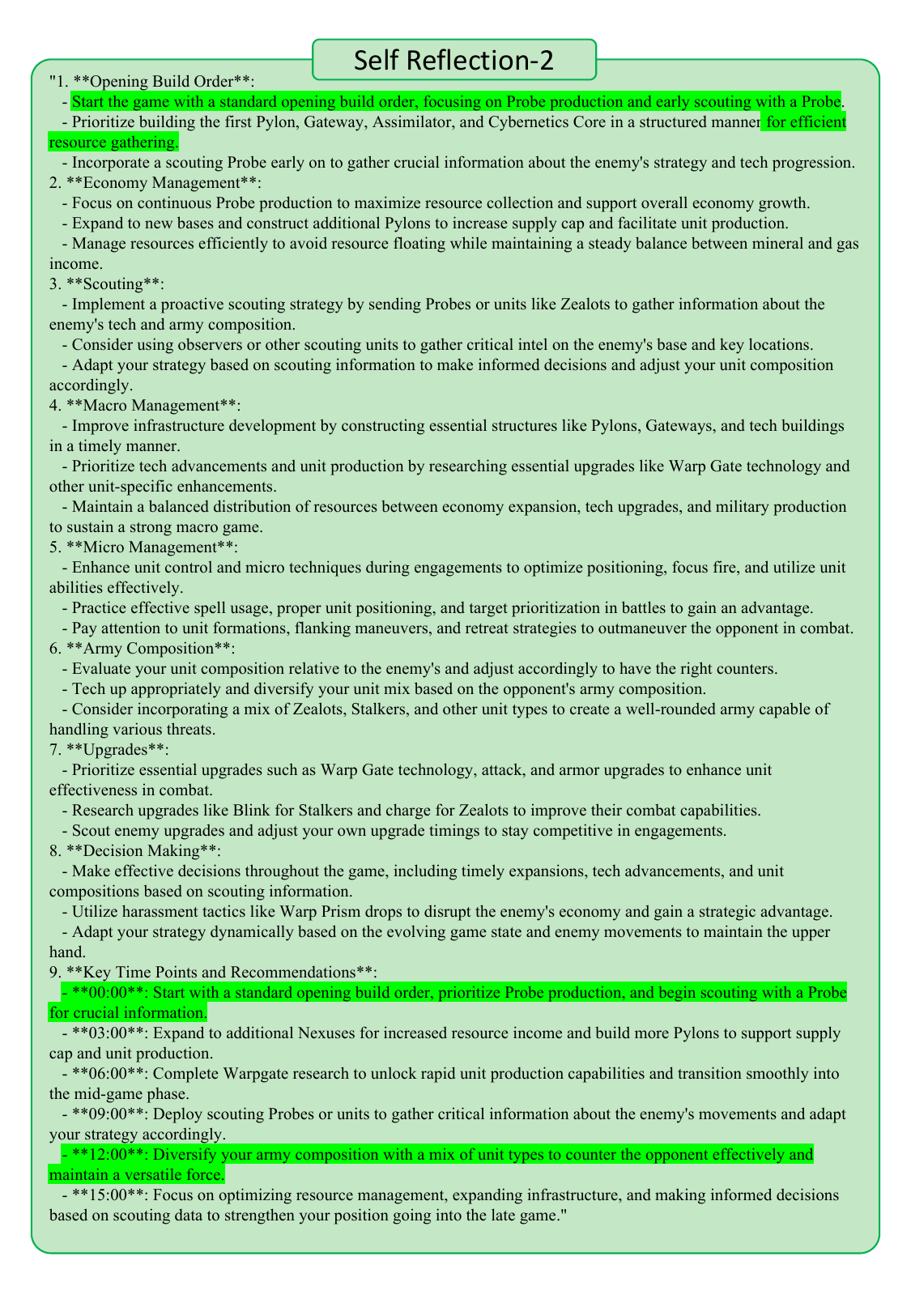}}
\caption{\textbf{Self Reflection2 with our method against Very Hard built-in AI}} 
\end{figure}


\begin{thebibliography}{8}


\bibitem{SC2LE}
Vinyals, O., et. al. Starcraft II: A new challenge for reinforcement learning. arxiv preprint, arxiv:1708.04782 (2017).
\bibitem{AlphaStar}
Vinyals, O., Babuschkin, I., et. al. AlphaStar: Mastering the real-time strategy game StarCraft II. DeepMind blog, 2, 20 (2019).
\bibitem{ChatGPT}
OpenAI ChatGPT team, \url{https://openai.com/chatgpt/} (2022)
\bibitem{GPT-4}
OpenAI GPT-4 team, \url{https://openai.com/index/gpt-4/} (2023)
\bibitem{Baseline}
Ma Weiyu, et. al. LLMs play StarCraft II: Benchmarks and a chain of summarization approach. arXiv preprint, arXiv:2312.11865 (2023).
\bibitem{VDN}
Peter Sunehag, et. al. Value-decomposition networks for cooperative multi-agent learning. arXiv preprint, arXiv: 1706.05296 (2017).
\bibitem{QMIX}
Rashid, Tabish, et. al. Monotonic value function factorisation for deep multi-agent reinforcement learning. Journal of Machine Learning Research 21.178: 1-51 (2020).
\bibitem{WQMIX}
Rashid, Tabish, et. al. Weighted Qmix: Expanding monotonic value function factorisation for deep multi-agent reinforcement learning. Advances in neural information processing systems 33: 10199-10210 (2020).
\bibitem{MAPPO}
Yu, Chao, et. al. The surprising effectiveness of ppo in cooperative multi-agent games. Advances in Neural Information Processing Systems 35: 24611-24624 (2022).
\bibitem{MADDPG}
Lowe, Ryan, et. al. Multi-agent actor-critic for mixed cooperative-competitive environments. Advances in neural information processing systems 30 (2017).
\bibitem{RLforSC}
Liu, Ruo-Ze, et. al. On efficient reinforcement learning for full-length game of starcraft II. Journal of Artificial Intelligence Research 75: 213-260 (2022).
\bibitem{Claude-2}
Anthropic Claude-2 team, \url{https://www.anthropic.com/news/claude-2} (2023).
\bibitem{LLAMA-3}
Meta Llama team, \url{https://github.com/meta-llama/llama3} (2024).
\bibitem{PaLM-2}
Anil, Rohan, et al. Palm 2 technical report. arXiv preprint, arXiv:2305.10403 (2023).
\bibitem{SwarmBrain}
Shao, X., Jiang, W., et. al. SwarmBrain: Embodied agent for real-time strategy game StarCraft II via LLMs. arXiv preprint, arXiv:2401.17749 (2024).
\bibitem{reflextion}
Shinn, Noah, et al. Reflexion: Language agents with verbal reinforcement learning. arXiv preprint, cs.AI/2303.11366 (2023).
\bibitem{Agent-Pro}
Zhang, Wenqi, et al. Agent-pro: Learning to evolve via policy-level reflection and optimization. arXiv preprint, arXiv:2402.17574 (2024).
\end{thebibliography}
\end{document}